\documentclass[11pt,twocolumn]{article}

\usepackage[margin=0.85in,columnsep=0.3in]{geometry}
\usepackage{setspace}
\usepackage{amsmath,amssymb,amsfonts}
\usepackage{amsthm}

\usepackage{graphicx}
\usepackage{booktabs}
\usepackage{tabularx}
\usepackage{multirow}
\usepackage{float}
\usepackage{xcolor}
\usepackage{tikz}
\usetikzlibrary{shapes,arrows,positioning,fit,backgrounds}

\usepackage{enumitem}
\setlist{nosep,leftmargin=*}

\newtheorem{definition}{Definition}

\usepackage[colorlinks=true,linkcolor=blue!60!black,citecolor=blue!60!black,urlcolor=blue!60!black]{hyperref}
\usepackage[round,authoryear]{natbib}

\usepackage{parskip}
\usepackage{titling}

\usepackage{titlesec}
\titleformat{\section}{\normalsize\bfseries}{\thesection}{1em}{}
\titleformat{\subsection}{\small\bfseries}{\thesubsection}{1em}{}
\titleformat{\subsubsection}{\small\itshape}{\thesubsubsection}{1em}{}
\titlespacing*{\section}{0pt}{0.9em}{0.4em}
\titlespacing*{\subsection}{0pt}{0.6em}{0.25em}

\pretitle{\begin{center}\Large\bfseries}
\posttitle{\end{center}\vspace{0.8em}}
\preauthor{\begin{center}\normalsize}
\postauthor{\end{center}\vspace{-0.5em}}
\predate{\begin{center}\small}
\postdate{\end{center}}

\title{The Keyhole Effect: Why Chat Interfaces Fail at Data Analysis\\[0.2em]
\large A Cognitive Science Framework for Hybrid Human-AI Interaction}

\author{%
\textbf{Mohan Reddy}\\[0.4em]
{\small Chief Scientist, Cornerstone AI Labs\\[0.2em]
\texttt{mreddy@csod.com}}%
}

\date{}

\begin{document}

\twocolumn[
\maketitle
\begin{@twocolumnfalse}
\begin{abstract}
\noindent Chat has become the default interface for AI-assisted data analysis. For multi-step, state-dependent analytical tasks, this is a mistake. Building on Woods' (1984) Keyhole Effect (the cognitive cost of viewing large information spaces through narrow viewports), I show that chat interfaces systematically degrade analytical performance through five mechanisms: (1) constant content displacement defeats hippocampal spatial memory systems; (2) hidden state variables exceed working memory capacity ($\sim$4 chunks under load); (3) forced verbalization triggers verbal overshadowing, degrading visual pattern recognition; (4) linear text streams block epistemic action and cognitive offloading; (5) serialization penalties scale with data dimensionality. I formalize cognitive overload as $O = \max(0, m - v - W)$ where $m$ is task-relevant items, $v$ is visible items, and $W$ is working memory capacity. When $O > 0$, error probability increases and analytical biases (anchoring, confirmation, change blindness) amplify. Eight hybrid design patterns address these failures: Generative UI, Infinite Canvas, Deictic Interaction, State Rail, Ghost Layers, Mise en Place, Semantic Zoom, and Probabilistic UI. Each pattern targets specific cognitive bottlenecks while preserving natural language for intent specification and synthesis. Well-scaffolded conversational systems that encode expert priors may reduce load for guided tasks; the framework applies most strongly to open-ended exploration. The paper concludes with falsifiable hypotheses and experimental paradigms for empirical validation.
\end{abstract}
\vspace{0.3em}
\vspace{0.8em}
\end{@twocolumnfalse}
]

%==============================================================================
\section{Introduction}
%==============================================================================

ChatGPT launched in November 2022 and suddenly conversation became the interface for everything: writing, coding, analysis, research. The implicit assumption is simple: if AI can understand natural language, natural language must be the optimal way to interact with AI.

This assumption ignores 40 years of human-computer interaction research and a century of cognitive psychology.

We are in the skeuomorphic phase of AI interaction. Early websites looked like newspaper broadsheets. Early e-books simulated page turns. New technologies routinely begin by imitating the forms they replace before discovering their native idiom. Chat-based AI imitates human conversation, a familiar and comfortable metaphor, but conversation is not the native idiom of data exploration. The mismatch is not aesthetic. It is cognitive.

The shift from command-line to graphical interfaces in the 1980s was not aesthetic. It was cognitive. \citet{shneiderman1983} showed that direct manipulation interfaces exploit human spatial cognition to reduce memory load. Recognition beats recall. Pointing beats typing. Seeing beats remembering. Chat-based AI reverses these gains.

I propose that chat works for two phases of analytical work: \textbf{stating intent} and \textbf{synthesizing results}. It fails at the three phases that constitute most of the actual work: \textbf{exploration} (foraging, filtering, hypothesis surfacing), \textbf{comparison} (side-by-side evaluation of alternatives), and \textbf{drill-down} (iterative refinement and provenance tracking).

The \emph{Keyhole Effect}, first identified by \citet{woods1984} in the context of process control displays, captures this failure. Woods observed that when virtual information spaces exceed the viewport capacity of displays, users can see only a small piece at a time, leading to navigation burdens and ``getting lost'' phenomena. \citet{bennett2012} later formalized \emph{visual momentum} as a design principle to counteract the effect through spatial dedication, perceptual landmarks, and display overlap.

Chat interfaces for data analysis represent an extreme case of the keyhole problem, and in some ways a regression. Unlike the display systems Woods analyzed, which at least permitted navigation and spatial organization, chat imposes a fundamentally linear structure that compounds the original problem: constant content displacement, forced verbalization of spatial information, and blocked epistemic action. You are trying to comprehend a vast, multi-dimensional dataset through a narrow scrolling window. You lose spatial context. You cannot compare side-by-side. Your previous findings disappear off the top of the screen. You forget what filters are active. You type when you should click.

This paper formalizes interface-cognition mismatch as an overload function, synthesizes six cognitive science frameworks (spatial cognition, working memory, dual coding, verbal overshadowing, epistemic action, distributed cognition) to explain why chat fails for analytical work, catalogs eight hybrid design patterns with cognitive justifications, and traces a morphic trajectory from current chat-with-artifacts through disposable just-in-time interfaces toward ambient object-oriented AI. The research agenda offers falsifiable hypotheses for empirical validation.

%==============================================================================
\section{Formalizing the Keyhole Effect}
\label{sec:formal}
%==============================================================================

Let an analysis moment require maintaining a set of \emph{simultaneously relevant items}: state variables (filters, time windows, cohorts), artifacts (charts, tables, model outputs), and hypotheses (candidate explanations, competing stories).

\begin{definition}[Cognitive Overload: Base Model]
Define:
\begin{itemize}
    \item $m$ = number of relevant items needed to reason correctly
    \item $v$ = number of items externally visible (persistent, glanceable)
    \item $W$ = working-memory capacity in chunks ($\approx 4$ under load) \citep{cowan2001}
\end{itemize}
The internal maintenance load is $L_{\text{internal}} = \max(0, m - v)$, and cognitive overload is:
\begin{equation}
\boxed{O = \max(0, m - v - W)}
\label{eq:overload}
\end{equation}
\end{definition}

This base model treats items as homogeneous and visibility as binary. Real systems are more complex: a pinned chart contributes more to $v$ than a minimized panel; a core hypothesis weighs more than a peripheral filter. An extended model would incorporate item weights $w_i$, graded visibility functions $\sigma_i \in [0,1]$ reflecting salience and accessibility, and dynamic $W$ influenced by expertise and chunking:
\begin{equation}
O_{\text{ext}} = \max\left(0, \sum_i w_i - \sum_i w_i \sigma_i - W(e, c)\right)
\label{eq:extended}
\end{equation}
where $e$ denotes expertise and $c$ captures external chunking aids. The base model (Equation~\ref{eq:overload}) is the degenerate case with $w_i = 1$ and $\sigma_i \in \{0,1\}$. I use the base model throughout for tractability, treating it as a first-order approximation that captures the qualitative dynamics while acknowledging that empirical work will require the extended formulation.

\textbf{Operationalizing $m$ and $v$.} For experimental purposes, $m$ can be measured as the count of: active filters, concurrent data views required for comparison, hypothesis cards under consideration, and state variables (time range, cohort definition, aggregation level). $v$ can be measured as the count of these items that are simultaneously visible without scrolling or retrieval actions. In pure chat, $v \approx 1$ (the current response); with a State Rail showing 4 active filters, $v = 5$; with a full dashboard, $v$ approaches $m$.

A candidate mapping from overload to error probability follows an exponential form:
\begin{equation}
P(\text{error}) \approx 1 - e^{-\lambda O}, \quad \lambda > 0
\label{eq:error}
\end{equation}
This functional form is motivated by analogy to signal detection under noise, where performance degrades exponentially with load, but should be treated as a hypothesis rather than a derived result. Alternative candidates include logistic functions or diffusion-to-bound models; the appropriate form and parameterization of $\lambda$ should be determined empirically.

Chat interfaces force $v \rightarrow 1$ (a single scrolling stream), while analysis drives $m$ upward, so $O$ grows predictably. Hybrid interfaces raise $v$ by externalizing state and artifacts, shrinking $O$.

\subsection{The Serialization Penalty}

Information is inherently dimensional; text is inherently serial. To convey a matrix of numbers, a network of relationships, or a spatial distribution through chat, you must \emph{serialize} it. This compression is lossy.

As a first approximation, define a \emph{Serialization Penalty} $S(d)$ that captures the cost of flattening $d$-dimensional structure into a one-dimensional stream:
\begin{equation}
S(d) = \alpha(d - 1), \quad d \geq 1, \; \alpha > 0
\end{equation}
This linear form is a simplification; actual costs likely depend on structure (sparse vs. dense), redundancy (compressible patterns), and prior knowledge (familiar schemas require less description). Information-theoretic arguments suggest $S(d)$ may scale with the entropy of the structure rather than raw dimensionality. The linear approximation captures the qualitative insight that serialization cost increases with structural complexity.

In a graphical interface, $d$-dimensional data can be presented directly: $S(1) = 0$. In chat, even a two-dimensional table requires row-by-row enumeration: $S(2) = \alpha$. For relational structures or multi-panel comparisons, $S$ scales with complexity.

Augmenting Equation~\ref{eq:overload}:
\begin{equation}
O' = O + S(d) = \max(0, m - v - W) + \alpha(d - 1)
\end{equation}
This explains why chat feels progressively more exhausting as analytical tasks become structurally richer. The penalty is not merely about \emph{volume} of information but about \emph{dimensionality} of structure.

\begin{table}[t]
\centering
\caption{Overload comparison across interface types for a typical analytical task with $m=8$ relevant items. Values are illustrative; empirical validation required.}
\label{tab:overload}
\small
\begin{tabular}{lccc}
\toprule
\textbf{Interface Type} & $v$ & $L_{\text{int}}$ & $O$ \\
\midrule
Chat-only & 1 & 7 & \textbf{3} \\
Chat + State Rail & 4 & 4 & 0 \\
Full Hybrid Dashboard & 7 & 1 & 0 \\
\bottomrule
\end{tabular}
\end{table}

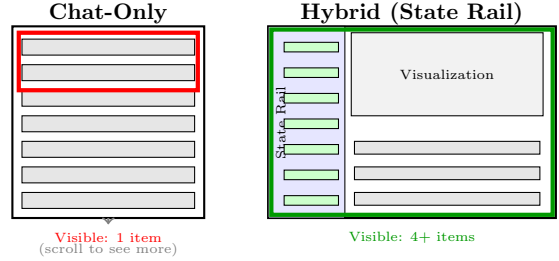
\begin{figure}[t]
\centering
\begin{tikzpicture}[scale=0.85, transform shape]
% Chat-only interface
\begin{scope}[shift={(0,0)}]
\node[font=\small\bfseries] at (1.5,3.2) {Chat-Only};
\draw[thick] (0,0) rectangle (3,3);
\foreach \y in {0.3,0.7,1.1,1.5,1.9,2.3,2.7} {
  \draw[fill=gray!20] (0.15,\y-0.15) rectangle (2.85,\y+0.1);
}
\draw[ultra thick, red] (0.1,2.0) rectangle (2.9,2.9);
\node[font=\tiny, red] at (1.5,-0.3) {Visible: 1 item};
\draw[->, thick, gray] (1.5,0) -- (1.5,-0.1);
\node[font=\tiny, gray] at (1.5,-0.5) {(scroll to see more)};
\end{scope}

% Hybrid interface
\begin{scope}[shift={(4,0)}]
\node[font=\small\bfseries] at (2.25,3.2) {Hybrid (State Rail)};
\draw[thick] (0,0) rectangle (4.5,3);
% State rail sidebar
\draw[fill=blue!10] (0,0) rectangle (1.2,3);
\node[font=\tiny, rotate=90] at (0.2,1.5) {State Rail};
\foreach \y in {0.3,0.7,1.1,1.5,1.9,2.3,2.7} {
  \draw[fill=green!20] (0.25,\y-0.1) rectangle (1.1,\y+0.05);
}
% Main area
\draw[fill=gray!10] (1.3,1.6) rectangle (4.3,2.9);
\node[font=\tiny] at (2.8,2.25) {Visualization};
\foreach \y in {0.3,0.7,1.1} {
  \draw[fill=gray!20] (1.35,\y-0.1) rectangle (4.25,\y+0.1);
}
\draw[ultra thick, green!60!black] (0.05,0.05) rectangle (4.45,2.95);
\node[font=\tiny, green!60!black] at (2.25,-0.3) {Visible: 4+ items};
\end{scope}
\end{tikzpicture}
\caption{The Keyhole Effect illustrated. \textbf{Left:} Chat-only interface shows one item at a time; users must scroll and remember. \textbf{Right:} Hybrid interface with State Rail externalizes filter state, raising $v$ and eliminating cognitive overload.}
\label{fig:keyhole}
\end{figure}

%==============================================================================
\section{The Neuroscience of Spatial Cognition}
%==============================================================================

\subsection{Place Cells and the Cognitive Map}

In 1971, John O'Keefe and Jonathan Dostrovsky made a discovery that would eventually earn O'Keefe the 2014 Nobel Prize in Physiology or Medicine. Recording from neurons in the hippocampus of freely moving rats, they found cells that fired only when the animal occupied specific locations in its environment. They called these ``place cells'' \citep{okeefe1971}.

The implications extended beyond navigation. The hippocampus was not merely involved in memory. It was constructing an internal spatial map of the environment.

O'Keefe and Lynn Nadel formalized this in their 1978 book \emph{The Hippocampus as a Cognitive Map} \citep{okeefe1978}. They proposed that the brain comes equipped to impose a three-dimensional Euclidean framework on experience. This is not learned. It is architecture. The hippocampus provides what they called an ``absolute, nonegocentric spatial framework'' that allows landmarks and goals to be encoded in terms of their allocentric (world-centered) locations.

Place cells have remarkable properties. Their firing patterns remain stable for months in constant environments. In one study, place fields persisted unchanged for 153 days \citep{thompson1990}. The brain maintains a durable spatial representation that can be accessed reliably over extended periods.

Later discoveries expanded this picture. The Mosers (who shared the 2014 Nobel with O'Keefe) found grid cells in the entorhinal cortex that fire in regular hexagonal patterns, providing a metric coordinate system \citep{nobel2014}. Head direction cells encode orientation. Border cells fire near environmental boundaries. Together, these cell types form a neural GPS.

But the cognitive map does more than navigation. It provides the substrate for episodic memory, planning, and imagination. When you remember where you put your keys, when you plan a route to work, when you imagine rearranging furniture, you are engaging hippocampal spatial circuits.

\subsection{Implications for Interface Design}

The existence of dedicated neural machinery for spatial representation has implications for interface design, though the connection should be framed as converging evidence rather than direct causal mechanism. The place cell and grid cell research establishes that the brain has specialized systems for encoding spatial relationships, distinct from the systems that process sequential verbal information.

The hypothesis is architectural: when you see a chart on the left side of a dashboard and a table on the right, you may be engaging spatial encoding systems that support durable, location-based retrieval. You know where things are. You can return to them. You can reason about relationships based on spatial proximity.

When you read a conversation transcript, you engage different systems. Sequential verbal processing runs through the phonological loop of working memory \citep{baddeley1974}. It is serial. It decays quickly. It has severe capacity limits.

Whether 2D screen layouts directly recruit hippocampal place cells (as opposed to engaging analogous spatial processing at the cognitive level) remains an open empirical question. The argument here is at the level of cognitive architecture: humans have distinct systems for spatial and verbal processing, and interfaces should allocate tasks to the appropriate system. Chat interfaces force spatial problems into sequential channels. This is a mismatch between task demands and cognitive architecture, regardless of the specific neural implementation.

\subsection{Spatial Memory is Hierarchical}

Research on spatial memory reveals a hierarchical structure that interfaces can either exploit or defeat. People remember the general layout of a space first, then use that layout to cue specific target locations.

\citet{cockburn2004} demonstrated that users develop spatial memory for interface locations and use it for rapid, recognition-based access. When interface elements maintain stable positions, users build cognitive maps of their digital workspace just as they build cognitive maps of physical environments.

Chat destroys this hierarchical structure. Each new message changes the spatial position of all previous content. There is no stable layout to learn. Users cannot develop ``that chart was in the upper left'' intuitions because by the next message, the chart has moved.

\subsection{The Memory Palace as Interface Metaphor}

The method of loci (``memory palace'') encodes items in imagined spatial locations and retrieves them by mentally walking a route \citep{bower1970,yates1966}. This is not a party trick; it exploits spatial scaffolding for durable recall.

A 2D analysis canvas is a shared external memory palace for data work: hypotheses live in places; comparisons become rooms; provenance is adjacency. Chat is an amnesic corridor: you can walk forward, but your landmarks slide behind you.

%==============================================================================
\section{Working Memory: The Bottleneck Chat Ignores}
%==============================================================================

\subsection{Miller's Chunks and Capacity Limits}

In 1956, George Miller published ``The Magical Number Seven, Plus or Minus Two,'' one of the most cited papers in psychology \citep{miller1956}. Miller observed that humans can hold approximately seven chunks of information in immediate memory.

The concept of ``chunk'' is key. A chess master looking at a board does not see 32 individual pieces. They see familiar configurations, each functioning as a single chunk. Expertise is partly the development of increasingly complex chunks that pack more information into fixed-capacity storage.

Subsequent research revised Miller's estimate downward. \citet{cowan2001} proposed that the true limit is closer to four chunks when rehearsal and long-term memory support are controlled. The key point is uncontroversial: working memory has hard capacity limits.

Chat interfaces make no accommodation for these limits. A complex analytical session might involve tracking multiple filters, several visualizations, comparison conditions, and intermediate findings. Each competes for the same limited working memory slots.

\subsection{Chunking and the External World}

Miller's insight about chunking has a corollary that interface designers have long understood: the environment can do the chunking for you. When information is organized spatially in persistent displays, users do not need to hold it all in working memory. They can offload storage to the interface.

A dashboard with six charts is not six chunks competing for working memory. It is six spatially distinct regions that can be accessed on demand through eye movements. The cognitive load is dramatically lower than maintaining six mental representations.

Chat provides no chunking support. Previous messages scroll away. The user must either remember their content or scroll back to retrieve it. Both options consume cognitive resources that could otherwise be devoted to analytical thinking.

\subsection{The Visuospatial Sketchpad}

Baddeley and Hitch's model of working memory distinguishes the visuospatial sketchpad from the phonological loop \citep{baddeley1974}. The sketchpad maintains visual and spatial information. The loop maintains verbal and acoustic information. These are separate systems with separate capacities.

Chat interfaces engage primarily the phonological loop: reading text, composing queries, parsing responses. The visuospatial sketchpad remains underutilized even when the task is spatial.

This is a waste of cognitive resources. The visuospatial system has its own capacity, separate from verbal working memory. Good interfaces engage both systems simultaneously. A user can look at a visualization (visuospatial) while formulating a verbal question about it. This is dual-channel processing, and it is more efficient than forcing everything through a single channel.

\citet{logie1986} demonstrated that unattended visual material has privileged access to spatial processing mechanisms. Visual representations are encoded automatically and rapidly in ways that verbal descriptions are not. When you see a chart, you immediately perceive patterns, outliers, trends. Describing those same patterns in text requires serial processing and conscious effort.

\subsection{Serial Position Effects}

Free recall shows consistent serial position effects: better memory for items near the beginning and end than those in the middle \citep{murdock1962}. A long chat session is literally a serial list of cognitive artifacts, promoting:
\begin{itemize}
    \item \textbf{Primacy bias}: early framing anchors later exploration
    \item \textbf{Recency bias}: the latest chart or explanation dominates
    \item \textbf{Middle amnesia}: intermediate caveats are forgotten
\end{itemize}
The result is coherence driven by narrative thread rather than careful comparison.

%==============================================================================
\section{Dual Coding: Why Pictures Beat Words}
%==============================================================================

\subsection{Paivio's Theory}

Allan Paivio's dual coding theory, developed in the 1960s and 1970s, proposes that the mind operates with two distinct classes of mental representation: verbal codes (logogens) and imagery codes (imagens) \citep{paivio1971,paivio1986}. These are processed by separate systems that can operate independently but also interact.

The key finding: concrete information that engages both systems is remembered better than abstract information that engages only the verbal system. This is why nouns like ``truck'' are easier to remember than nouns like ``truth.'' The concrete word activates both a verbal representation and a mental image.

Paivio demonstrated the \emph{picture superiority effect}: when people study lists containing both pictures and words, they recall twice as many pictures as words. Images receive automatic dual coding. Words may or may not evoke imagery depending on their concreteness.

\subsection{Verbal Overshadowing: When Language Degrades Vision}

A more troubling finding emerges from Jonathan Schooler's research on \emph{verbal overshadowing} \citep{schooler1990}. Participants who verbally described a face after viewing it were \emph{less} accurate at recognizing that face later than participants who performed an unrelated task. The act of putting visual experience into words actively degraded the visual memory.

This finding has been extensively replicated. A Registered Replication Report involving multiple independent laboratories confirmed the core phenomenon \citep{alogna2014}. However, boundary conditions matter: the effect is strongest when verbalization occurs immediately after encoding, when test stimuli are similar to targets (requiring fine discrimination), and when the visual information is holistic rather than featural \citep{meissner2001}. The effect may be attenuated for highly distinctive stimuli or when participants are warned about potential interference.

Despite these boundary conditions, the core implication for interface design holds: forcing users to verbally describe visual patterns (as chat interfaces require) risks degrading those visual representations. The verbal code is a lossy compression. When you describe a chart's ``upward trend with a spike around May,'' you are translating a precise visual signal into an imprecise semantic approximation.

Chat interfaces demand this translation constantly. ``Show me the outlier in the top right'' requires transcoding from a visual percept to a verbal description. The transcoding is cognitively expensive (consuming working memory for the translation) and representationally lossy (the verbal label lacks the precision of the visual encoding). The verbal description may interfere with subsequent visual recognition of the same pattern.

The implication: chat interfaces do not merely hide data; they may actively degrade the user's mental model by forcing verbalization of non-verbal structure.

\subsection{Implications for Data Interfaces}

Data analysis is about patterns, relationships, and structures. These are spatial concepts that benefit from visual representation. A scatterplot showing customer segments is processed differently than a verbal description of those same segments. The visual version engages imagery systems. It is encoded more durably. It is retrieved more easily.

Chat interfaces default to verbal output. Even when they include images, those images are embedded in a text stream that the user must scroll to access. Chat-based analysis therefore receives weaker encoding than equivalent analysis performed through visual interfaces. The findings are less memorable. The patterns are less salient.

\subsection{Sequential vs. Parallel Processing}

Verbal representations are processed sequentially. You read a sentence word by word. You cannot take in a paragraph at a glance.

Visual representations are processed in parallel. You see a chart and immediately perceive the overall shape, the distribution, the outliers. Details emerge with focused attention, but the gestalt is available instantly.

Analytical insight often comes from perceiving relationships between multiple data points, precisely what parallel visual processing captures. When you force analysis through sequential verbal channels, you fragment the gestalt into pieces that must be laboriously reassembled in working memory.

%==============================================================================
\section{Distributed Cognition}
%==============================================================================

\subsection{Cognition in the Wild}

Edwin Hutchins, working at UC San Diego, developed the theory of distributed cognition through ethnographic studies of navigation teams aboard Navy ships \citep{hutchins1995}. His central insight: cognitive processes are not confined to individual brains. They are distributed across people, tools, and environments.

In Hutchins' famous example, a ship's navigation system includes the captain, the navigator, various crew members, charts, instruments, and procedures. No single person holds the complete representation needed to navigate safely. Knowledge is distributed across the system. The ``mind'' doing the navigation is the entire sociotechnical assembly.

\subsection{Cognitive Offloading}

A key concept from distributed cognition is \emph{cognitive offloading}: using external resources to reduce internal memory and processing demands. When you write a shopping list, you offload memory. When you use a calculator, you offload arithmetic. When you arrange papers on your desk to represent project status, you offload planning.

Well-designed interfaces enable cognitive offloading. Dashboards externalize data relationships. State indicators externalize filter settings. Spatial layouts externalize analytical structure. Each externalization reduces the cognitive burden on the user.

Chat interfaces are poor at cognitive offloading. The conversation is not a stable external representation. It is a transient stream. To offload to it, you would need to scroll back and search. The cost of retrieval undermines the benefit of externalization.

\citet{kirshmaglio1994} demonstrated that expert Tetris players physically rotate falling pieces to reduce the mental effort of imagining rotations. They offload spatial reasoning onto the interface. Chat interfaces offer no equivalent affordance for analytical tasks. There is nothing to manipulate, nothing to arrange, nothing to externalize analytical state onto.

\subsection{Epistemic Action: Thinking with Your Hands}

Kirsh and Maglio's Tetris work reveals a deeper distinction between \emph{pragmatic action} and \emph{epistemic action}. Pragmatic actions move you closer to a goal directly: typing a query, clicking submit. Epistemic actions change the world to make it easier to think: shuffling Scrabble tiles to trigger word recognition, rotating a map to align with your heading, spreading papers across a desk to see relationships.

Epistemic actions are how humans extend cognition into the environment. They transform internal computation into external manipulation. They are not incidental to thought; they are constitutive of it.

Chat blocks epistemic action. You cannot shuffle the AI's response. You cannot group related findings spatially. You cannot stack alternatives for comparison. You cannot rotate a visualization to see it differently. The medium resists manipulation. You are forced to think \emph{at} the interface rather than \emph{with} it.

The Keyhole Effect is therefore not just a visibility problem. It is an \emph{agency} problem. Chat interfaces prevent users from thinking with their hands.

\subsection{The Interface as Cognitive Partner}

Hutchins argued that we should analyze cognitive systems, not individual minds. A pilot and their cockpit form a cognitive system. The boundaries of the system are determined by the flow of information, not by skin and skull \citep{hollan2000}.

This perspective reframes interface design. The question is not ``how do we present information to the user?'' The question is ``how do we build a cognitive system that includes the user?''

Chat interfaces are poor cognitive partners for analytical tasks. They do not maintain state. They do not represent spatial relationships. They do not support parallel comparison. They treat the user as an isolated processor who must hold everything in their head.

%==============================================================================
\section{Interaction Speed and Language Friction}
%==============================================================================

\subsection{Fitts' Law and Pointing Speed}

Fitts' Law models human movement in targeting tasks \citep{fitts1954}:
\begin{equation}
MT = a + b \log_2\!\left(\frac{D}{W} + 1\right)
\end{equation}
where $MT$ is movement time, $D$ is distance, and $W$ is target width. \citet{card1978} applied this to computer input devices. Mouse throughput ranges from 4 to 10 bits per second. Target acquisition takes hundreds of milliseconds. Clicking a clearly labeled button is fast.

\subsection{The Typing Bottleneck}

Skilled typists manage 40-80 words per minute. But formulating an analytical command requires more than raw typing. Compose the query. Determine phrasing. Type it. Revise for clarity. A simple refinement like ``filter for enterprise users'' takes 5-10 seconds.

Clicking a checkbox labeled ``Enterprise'' takes 500 milliseconds.

That is an order of magnitude difference. Across 50 refinement operations in an analytical session, the cumulative time cost becomes substantial. Worse, each typed command interrupts analytical flow. You context-switch from thinking about data to thinking about words.

\begin{table}[t]
\centering
\caption{Estimated interaction time comparison across modalities (based on Fitts' Law and typing speed models; empirical validation needed).}
\label{tab:time}
\small
\begin{tabular}{lcc}
\toprule
\textbf{Action} & \textbf{Chat (s)} & \textbf{GUI (s)} \\
\midrule
Simple filter & 5.2 & 0.5 \\
Date range selection & 8.5 & 1.2 \\
Multi-select dropdown & 12.3 & 2.1 \\
Drill-down navigation & 7.8 & 0.8 \\
Compare views & 15.2 & 1.5 \\
\midrule
\textbf{50 operations} & \textbf{$\sim$8 min} & \textbf{$\sim$1 min} \\
\bottomrule
\end{tabular}
\end{table}

\subsection{The Ambiguity Problem}

Natural language is ambiguous by design. ``What is that outlier in the top right quadrant?'' relies on spatial deixis (\emph{this}, \emph{that}, \emph{here}). In person, you point. In a GUI, you click.

Chat lacks deictic capacity. You must convert pointing into description: ``the data point with approximately 87\% retention and \$450K revenue that appears separated from the main cluster.'' This is slow, error-prone, and may be interpreted differently than intended.

%==============================================================================
\section{Cognitive Load and Psychological Biases}
%==============================================================================

\subsection{Sweller's Framework}

Cognitive load theory distinguishes intrinsic load (inherent to the task), extraneous load (imposed by poor interface design), and germane load (devoted to schema construction and learning) \citep{sweller1988}. The goal is to minimize extraneous load so that cognitive resources can be devoted to intrinsic and germane processing.

Chat interfaces impose extraneous load on analytical tasks: formulating linguistic queries when a click would suffice, scrolling to find previous results, mentally tracking hidden state, reconstructing spatial relationships from verbal descriptions. None of this advances the analytical task. It is overhead.

\subsection{Empirical Evidence}

\citet{nguyen2022} ran a controlled experiment comparing chatbots to menu-based interfaces for identical tasks. The results were unambiguous. Chatbots produced lower perceived autonomy and higher cognitive load. Users reported that despite the ``naturalness'' of conversation, chatbots demanded more mental effort. The culprit: uncertainty. Open-ended conversation requires users to formulate precise prompts without visual scaffolding.

This finding aligns with theoretical predictions. Recognition is cognitively cheaper than recall. GUIs support recognition. Chat demands recall.

\subsection{Bias Amplification}

The Keyhole Effect is also a \emph{judgment} problem: interface shape changes which hypotheses feel plausible.

\paragraph{Anchoring.} Early fluent explanations become anchors that subsequent evidence must overcome \citep{tverskykahneman1974}. Chat makes this costly because the user must retrieve and compare sequential fragments.

\paragraph{Confirmation bias.} Hypothesis testing often becomes hypothesis-protecting: users request elaborations that confirm rather than disconfirm \citep{wason1960}. Interfaces that externalize competing hypotheses make disconfirmation cheap.

\paragraph{Illusion of explanatory depth.} People routinely overestimate how well they understand causal systems; fluent explanations can inflate this illusion without forcing operationalization \citep{rozenblitkeil2002}.

\paragraph{Change blindness.} Comparing two similar text outputs in a scroll-based transcript disrupts attention and invites missed differences \citep{rensink1997,simonschabris1999}.

%==============================================================================
\section{Eight Patterns for Hybrid Interfaces}
%==============================================================================

The solution is not abandoning conversational AI. It is recognizing where conversation belongs in the interaction ecology.

\subsection{Generative UI: The Ephemeral Application}

Instead of responding with text or static images, have the AI generate interactive software on demand. The query ``Analyze the churn spike in May'' produces a dashboard widget with a chart plus relevant controls: a churn probability slider, a region dropdown, date range selectors. You state intent through conversation. You refine through direct manipulation.

This engages both coding systems. The verbal query triggers the visual display. The visual display provides stable spatial structure for the visuospatial sketchpad. The user can offload state onto the interface. The cognitive load plummets.

Recent implementations demonstrate feasibility. Google's generative UI system produces complete interactive experiences from arbitrary prompts, with human evaluators preferring these outputs over standard markdown \citep{leviathan2025}. Vercel's AI SDK provides infrastructure for React-based generative interfaces, and LangChain offers similar capabilities through its generative UI framework. These systems show that the pattern is not merely theoretical but implementable at scale.

\subsection{The Infinite Canvas: Spatial Memory for Analysis}

Replace the linear chat feed with a two-dimensional workspace. You ask a question. A card with a visualization appears. You drag an arrow from that card to empty space and type ``Break this down by product.'' A new connected card appears. The spatial arrangement encodes analytical relationships that sequential conversation loses.

This exploits the hierarchical structure of spatial memory. The general layout provides context. Individual cards can be located within that layout. The user builds a cognitive map of their analysis.

\subsection{Deictic Interaction: Point-and-Ask}

Combine the precision of direct manipulation with the expressiveness of natural language. Hold a key to indicate pointing mode. Circle a cluster on a scatter plot. Ask ``What do these customers have in common?'' The system receives both the selection (specific data point IDs) and the query (analytical intent). No verbose spatial descriptions required.

This is how human communication works. We point and speak simultaneously.

\subsection{The State Rail: Persistent Context Display}

Hidden state kills analytical accuracy. Ten messages ago you filtered to Europe. Fifteen messages ago you excluded a product category. Both filters remain active but invisible.

A State Rail displays current analytical state in a persistent sidebar. Every filter, grouping, date range, and exclusion appears as a visible tag. The rail updates as you converse. Click the X on any tag to remove that constraint directly. This externalizes state and enables cognitive offloading. This raises $v$ in Equation~\ref{eq:overload}.

\subsection{Ghost Layers: Proactive Guidance}

Users face blank-slate paralysis. They see a chart but lack the intuition to formulate the right question.

Ghost layers invert the query-response dynamic. The AI analyzes each visualization and overlays subtle markers on statistically interesting features. A small indicator appears on an unusual revenue dip. Hover to see context. Click to drill down automatically. This reduces the recall burden. Recognition over recall, applied to analytical strategy.

\subsection{Mise en Place: The Prepared Workspace}

Chat establishes a ``fetch quest'' dynamic: the user requests, the AI retrieves, the user requests again. Each cycle is a round trip through the keyhole.

The \emph{Mise en Place} protocol inverts this. You state your analytical intent: ``I want to analyze customer churn.'' The AI does not respond with an answer. Instead, it \emph{generates a workspace}: the four most relevant visualizations arranged spatially, the three most likely filters pre-configured, anomalies already highlighted. The sous-chef prepares the ingredients on the counter; the analyst does the cooking.

This shifts the interaction from sequential retrieval to parallel exploration. The user surveys the prepared workspace, notices unexpected patterns, and begins manipulating directly. Intent through conversation; refinement through spatial engagement.

\subsection{Semantic Zoom: Resolution as Interaction}

The Keyhole Effect is partly a resolution problem. At any moment, you see either the summary or the detail, never both in relationship.

Semantic zoom redefines scrolling. The mouse wheel does not move the viewport; it changes the \emph{level of abstraction}. Scroll up: the chart collapses into a summary sentence. Scroll to center: the sentence expands into a dashboard. Scroll down: the dashboard bars explode into underlying data rows.

Each zoom level is a complete, coherent representation. The user navigates a hierarchy of abstraction rather than a stack of messages. Context is never lost because higher levels remain cognitively accessible through a single gesture.

\subsection{Probabilistic UI: Uncertainty Made Visible}

Traditional GUIs imply certainty. A toggle is on or off. A filter is applied or not. This works when the system knows exactly what the user wants.

LLMs are probabilistic. They \emph{infer} intent. They \emph{guess} which filter you meant. The deterministic visual language of conventional interfaces cannot express this uncertainty.

Probabilistic UI elements represent confidence visually. A filter tag with a glowing border says ``I inferred this, but I need verification.'' A partially transparent panel says ``This interpretation is uncertain.'' A ghost duplicate shows the alternative interpretation the system considered.

This bridges the gap between deterministic software and probabilistic AI. The user sees not just what the system decided, but how confident it is. Verification becomes recognition rather than reconstruction.

A design concern: surfacing uncertainty may increase cognitive load if every element displays confidence levels. Effective implementations should threshold visibility (only show uncertainty markers above/below meaningful confidence bounds), use progressive disclosure (details on hover/demand), and establish consistent visual language (e.g., saturation for confidence, border treatment for verification needs). The goal is legibility without overwhelm: make uncertainty visible when it matters for user decisions, invisible when the system is confident.

\begin{table}[t]
\centering
\caption{Cognitive benefits of hybrid patterns.}
\label{tab:patterns}
\small
\begin{tabularx}{\columnwidth}{lX}
\toprule
\textbf{Pattern} & \textbf{Cognitive Benefit} \\
\midrule
Generative UI & Dual coding; cognitive offloading \\
Infinite Canvas & Spatial memory; hierarchical recall \\
Deictic Interaction & Reduces $m$ (no verbal reconstruction) \\
State Rail & Raises $v$ (state visibility) \\
Ghost Layers & Recognition over recall \\
Mise en Place & Parallel exploration; reduced round trips \\
Semantic Zoom & Resolution hierarchy; context preservation \\
Probabilistic UI & Uncertainty legible; verification by recognition \\
\bottomrule
\end{tabularx}
\end{table}

\begin{figure*}[t]
\centering
\begin{tikzpicture}[scale=0.75, transform shape]
% Pattern 1: Generative UI
\begin{scope}[shift={(0,0)}]
\node[font=\small\bfseries] at (2,4.3) {Generative UI};
\draw[thick, rounded corners] (0,0) rectangle (4,4);
% Chat input
\draw[fill=blue!10, rounded corners] (0.2,0.2) rectangle (3.8,0.8);
\node[font=\tiny] at (2,0.5) {``Show churn by region''};
% Generated dashboard
\draw[fill=green!10, rounded corners] (0.2,1) rectangle (3.8,3.8);
\draw[fill=white] (0.4,2.2) rectangle (1.8,3.6);
\draw[thick] (0.5,2.4) -- (0.8,3.0) -- (1.1,2.8) -- (1.4,3.2) -- (1.7,2.9);
\draw[fill=white] (2,2.2) rectangle (3.6,3.6);
\foreach \x in {2.2,2.6,3.0,3.4} {
  \draw[fill=blue!40] (\x,2.4) rectangle (\x+0.25,2.4+0.8*rnd+0.3);
}
\draw[fill=gray!20, rounded corners] (0.4,1.2) rectangle (1.5,2);
\node[font=\tiny] at (0.95,1.6) {Filters};
\draw[fill=gray!20, rounded corners] (1.7,1.2) rectangle (3.6,2);
\node[font=\tiny] at (2.65,1.6) {Controls};
\end{scope}

% Pattern 2: Infinite Canvas
\begin{scope}[shift={(5,0)}]
\node[font=\small\bfseries] at (2,4.3) {Infinite Canvas};
\draw[thick, rounded corners] (0,0) rectangle (4,4);
\draw[fill=yellow!10] (0,0) rectangle (4,4);
% Cards with connections
\draw[fill=white, rounded corners] (0.3,2.5) rectangle (1.5,3.7);
\node[font=\tiny] at (0.9,3.1) {Query 1};
\draw[fill=white, rounded corners] (2,2.8) rectangle (3.2,3.8);
\node[font=\tiny] at (2.6,3.3) {Result};
\draw[fill=white, rounded corners] (2.2,1.2) rectangle (3.5,2.4);
\node[font=\tiny] at (2.85,1.8) {Drill-down};
\draw[fill=white, rounded corners] (0.5,0.5) rectangle (1.8,1.8);
\node[font=\tiny] at (1.15,1.15) {Compare};
% Arrows
\draw[->, thick, gray] (1.5,3.1) -- (2,3.3);
\draw[->, thick, gray] (2.6,2.8) -- (2.6,2.4);
\draw[->, thick, gray] (2.2,1.8) -- (1.8,1.5);
\end{scope}

% Pattern 3: State Rail
\begin{scope}[shift={(10,0)}]
\node[font=\small\bfseries] at (2,4.3) {State Rail};
\draw[thick, rounded corners] (0,0) rectangle (4,4);
% Sidebar
\draw[fill=blue!15] (0,0) rectangle (1.2,4);
\node[font=\tiny, rotate=90] at (0.15,2) {Active Filters};
\foreach \y/\t in {3.5/Region: EU, 3.0/Date: Q3, 2.5/Cohort: New, 2.0/Product: Pro} {
  \draw[fill=white, rounded corners] (0.15,\y-0.2) rectangle (1.1,\y+0.15);
  \node[font=\tiny, right] at (0.18,\y) {\t};
  \node[font=\tiny, red] at (1.0,\y) {$\times$};
}
% Main area
\draw[fill=gray!10] (1.3,0.2) rectangle (3.8,3.8);
\draw[fill=white] (1.5,1.8) rectangle (3.6,3.6);
\node[font=\tiny] at (2.55,2.7) {Visualization};
\draw[fill=gray!20, rounded corners] (1.5,0.4) rectangle (3.6,1.6);
\node[font=\tiny] at (2.55,1) {Chat};
\end{scope}

% Pattern 4: Deictic Interaction
\begin{scope}[shift={(15,0)}]
\node[font=\small\bfseries] at (2,4.3) {Deictic (Point-and-Ask)};
\draw[thick, rounded corners] (0,0) rectangle (4,4);
% Scatterplot
\draw[fill=gray!10] (0.2,1) rectangle (3.8,3.8);
\draw[->] (0.4,1.2) -- (0.4,3.6);
\draw[->] (0.4,1.2) -- (3.6,1.2);
% Points
\foreach \x/\y in {1/2,1.5/2.3,2/2.8,2.5/1.8,3/3.2,1.2/3,2.8/2.5} {
  \fill[blue!60] (\x,\y) circle (0.08);
}
% Circled selection
\draw[red, thick, dashed] (2.7,2.3) circle (0.5);
\fill[red] (2.8,2.5) circle (0.08);
\fill[red] (3,3.2) circle (0.08);
% Query
\draw[fill=blue!10, rounded corners] (0.2,0.2) rectangle (3.8,0.8);
\node[font=\tiny] at (2,0.5) {``What do these have in common?''};
\draw[->, thick, red] (2.7,1.8) -- (2.5,0.85);
\end{scope}
\end{tikzpicture}
\caption{Four hybrid design patterns. \textbf{Generative UI:} Chat input produces interactive dashboards. \textbf{Infinite Canvas:} 2D workspace preserves spatial relationships between queries. \textbf{State Rail:} Persistent sidebar externalizes filter state. \textbf{Deictic Interaction:} Circle data points and ask questions about the selection.}
\label{fig:patterns}
\end{figure*}
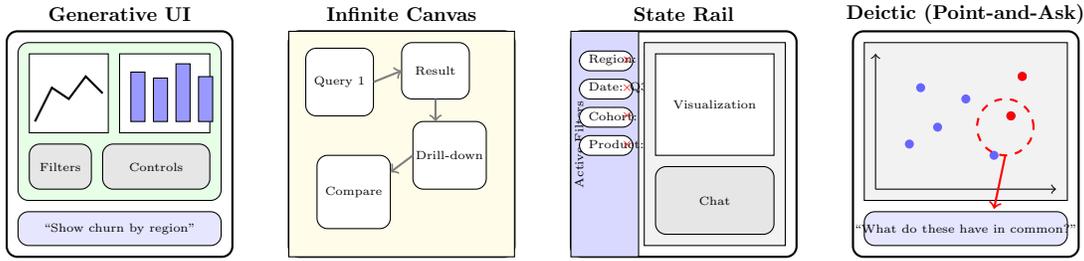

%==============================================================================
\section{The Pareto Principle of Analytical Interfaces}
%==============================================================================

These patterns suggest a practical heuristic.

\textbf{Chat handles 20\% of interactions:}
\begin{itemize}
    \item Initial intent specification (``Show me customer churn over time'')
    \item Complex one-off queries (``Compare retention rates for customers who used feature X within their first week versus those who discovered it later'')
    \item Final synthesis (``Summarize the key findings'')
\end{itemize}

\textbf{GUI handles 80\% of interactions:}
\begin{itemize}
    \item All refinement: filtering, sorting, grouping, drilling down
    \item All comparison: viewing alternatives side-by-side
    \item State management: understanding and modifying active filters
    \item Iterative exploration: following chains of related questions
\end{itemize}

Using Equation~\ref{eq:overload}: if $O=0$, chat-only is tolerable (simple tasks, small state); if $O \geq 1$, externalize state (raise $v$); if $O \gg 1$, chat-only becomes predictably error-prone.

The implication: the best chat interface for data analysis is a dynamic dashboard builder controlled by voice and text. Chat is the launcher. The generated GUI is the vehicle.

%==============================================================================
\section{The Morphic Trajectory: From Chat to Ambient Intelligence}
%==============================================================================

\begin{figure}[t]
\centering
\begin{tikzpicture}[scale=0.8, transform shape, 
    box/.style={draw, rounded corners, minimum width=2.2cm, minimum height=1.2cm, align=center, font=\tiny},
    arrow/.style={->, thick, >=stealth}]
    
% Phase 1
\node[box, fill=red!20] (p1) at (0,0) {Phase 1\\Chat +\\Artifact};
\node[font=\tiny, below=0.1cm of p1, text width=2cm, align=center] {Artifact is\\subordinate display};

% Phase 2
\node[box, fill=yellow!30] (p2) at (4,0) {Phase 2\\Just-in-Time\\GUI};
\node[font=\tiny, below=0.1cm of p2, text width=2cm, align=center] {Interface morphs\\to match intent};

% Phase 3
\node[box, fill=green!30] (p3) at (8,0) {Phase 3\\Object-Oriented\\AI};
\node[font=\tiny, below=0.1cm of p3, text width=2cm, align=center] {Data explains\\itself};

% Arrows
\draw[arrow] (p1) -- (p2);
\draw[arrow] (p2) -- (p3);

% Labels above
\node[font=\tiny\itshape] at (2,0.9) {disposable};
\node[font=\tiny\itshape] at (2,0.6) {software};
\node[font=\tiny\itshape] at (6,0.9) {ambient};
\node[font=\tiny\itshape] at (6,0.6) {intelligence};

% Keyhole indicator
\draw[thick, red!70] (0,-1.8) -- (0,-2.2);
\draw[thick, red!70] (-0.3,-2) -- (0.3,-2);
\node[font=\tiny, red!70] at (0,-2.5) {Keyhole};
\draw[thick, yellow!70!black] (4,-1.8) -- (4,-2.2);
\draw[thick, yellow!70!black] (3.5,-2) -- (4.5,-2);
\node[font=\tiny, yellow!70!black] at (4,-2.5) {Window};
\draw[thick, green!50!black] (8,-1.8) -- (8,-2.2);
\draw[thick, green!50!black] (7,-2) -- (9,-2);
\node[font=\tiny, green!50!black] at (8,-2.5) {Transparent};

\end{tikzpicture}
\caption{The morphic trajectory from chat-with-artifacts to ambient intelligence. The keyhole progressively widens until it disappears.}
\label{fig:trajectory}
\end{figure}
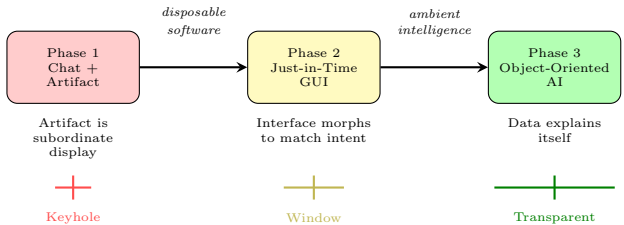

The patterns described above are not endpoints. They are waypoints on a trajectory away from the chat paradigm. Understanding this trajectory helps predict where interface design is heading and why. Figure~\ref{fig:trajectory} illustrates the three phases.

\subsection{Phase One: The Artifact}

Current implementations place chat and visual output side by side. Claude generates an artifact in a panel; the user converses in the chat window. This is progress: the visual output is persistent, interactive, and spatially distinct from the conversation.

But the artifact remains subordinate. To modify it, you must return to chat. The visual panel is a \emph{display}; it is not yet a \emph{workspace}. The user still operates through the keyhole when refinement is needed.

\subsection{Phase Two: Disposable Software}

The next phase dissolves the artifact into the interface itself. The ``application'' becomes a database of capabilities plus a set of data connections. The \emph{interface} is generated entirely in real-time based on context and intent.

You mention time-series data; the software becomes a timeline tool. You mention geography; it morphs into a map. You mention comparison; it generates a side-by-side layout. There is no ``Map Tab'' or ``Chart View.'' The system constructs the appropriate interface for that moment of work, then dissolves it when the task changes.

This is \emph{Just-in-Time GUI}: software that assembles itself around your needs rather than forcing your needs into predetermined forms. The chat window shrinks to a command line or disappears entirely, replaced by direct manipulation of the generated interface.

\subsection{Phase Three: Object-Oriented AI}

The terminal phase inverts the relationship between intelligence and data entirely. The AI is no longer a separate entity you converse with. The AI becomes a \emph{property of the data itself}.

You do not open a chat window. You open a dataset. You right-click a column, and the column speaks: ``I have 40\% missing values, concentrated in EU records from Q2.'' You select a cluster of points, and the cluster explains itself: ``These customers share high engagement but low conversion; the common factor is mobile-only access.''

Intelligence becomes ambient. Every UI element can answer questions about itself. The ``chat interface'' dissolves because there is nothing to chat \emph{with}. There is only data that knows itself and can explain itself on demand.

This is not science fiction. It is the logical endpoint of treating language models as general-purpose reasoning engines rather than conversational partners. The keyhole disappears because there is no longer a wall between you and the information. The room becomes transparent.

%==============================================================================
\section{Research Agenda}
\label{sec:agenda}
%==============================================================================

\subsection{Falsifiable Hypotheses}

\begin{description}[style=nextline,leftmargin=0pt,labelwidth=0pt,labelsep=0.6em]
    \item[H1 (Load):] For multi-step analytic tasks with hidden state, chat-only interfaces yield higher cognitive load than hybrids with persistent state and interactive views.
    \item[H2 (Accuracy):] Chat-only analysis produces more state-consistency errors (conclusions drawn under forgotten filters) than interfaces with state rails.
    \item[H3 (Bias):] Sequential chat increases anchoring and confirmation effects relative to interfaces enabling side-by-side counter-hypothesis comparison.
    \item[H4 (Comparison):] Side-by-side visualization reduces missed differences (change blindness) versus scroll-based transcript comparison.
\end{description}

\subsection{Experimental Paradigms}

\paragraph{Within-subject UI comparison.} Conditions: chat-only vs.\ hybrid (chat + state rail + interactive views). Measures: task time, accuracy, NASA-TLX, backtrack count, forgotten-filter errors. Analysis: mixed-effects models with participant as random effect.

\paragraph{Anchoring manipulation.} Provide an initial plausible explanation; measure persistence under contradictory evidence across UI conditions.

\paragraph{Deictic efficiency study.} Compare ``describe the point'' vs.\ ``point-and-ask'' for speed and accuracy on outlier identification and subgroup characterization.

%==============================================================================
\section{Discussion}
\label{sec:discussion}
%==============================================================================

\subsection{Relationship to Prior Work}

The framework developed here extends and synthesizes several research traditions. Woods' original keyhole effect \citeyearpar{woods1984} addressed navigation in complex display systems; I have argued that LLM chat interfaces represent an extreme and in some ways regressive case. Where Woods' operators could at least navigate between views, chat users face constant content displacement with no stable spatial structure to navigate.

\subsection{Hybrid and Decomposition-Based Systems}

Recent work demonstrates a continuum of approaches between pure chat and full GUI, several of which partially address the concerns raised here. Phasewise and Stepwise decomposition UIs externalize assumptions, plans, and code within a chat-centered container via progressive disclosure, achieving some benefits of state externalization without requiring full generative GUI. InsightLens shows that automatic extraction and multi-view organization of insights (via Insight Gallery, Minimap, Topic Canvas) can mitigate the keyhole effect within chat transcripts by raising $v$ through auxiliary panes. Talk2Data exemplifies routing between narration and executable code, returning figures and tables in ways congruent with the ``chat for intent, GUI for refinement'' notion proposed here.

ProactiveVA resonates with the Ghost Layers and Mise en Place patterns via proactive, context-aware suggestions and UI annotations. DASHBOARD QA highlights the difficulty of multi-step GUI reasoning and state traversal, reinforcing the need for explicit state rails and plan-tracking.

These systems suggest that the pure chat vs. hybrid distinction is better understood as a spectrum. Decomposition UIs that externalize editable assumptions within chat containers represent intermediate designs that may reduce $O$ without fully ``raising $v$'' in the spatial sense.

\subsection{When Guided Conversation Works: The StatZ Counterexample}

StatZ reports strong accuracy and time advantages of a guided conversational agent over mature GUI statistics tools, particularly for novices. This finding merits careful consideration, as it appears to challenge the central thesis.

The reconciliation lies in recognizing what StatZ's conversational agent provides: expert priors encoded in the system, guardrails that prevent common errors, and targeted suggestions that reduce the hypothesis space. In the terminology of this paper, StatZ \emph{raises effective $v$} not through spatial persistence but through \emph{conversational scaffolding} that externalizes expert knowledge into the dialogue structure. The user does not need to maintain statistical decision trees in working memory because the agent surfaces relevant options.

This suggests important boundary conditions: (1) for tasks where expert priors can be encoded and surfaced conversationally, guided chat may reduce load relative to complex GUIs that require users to discover functionality; (2) novices facing unfamiliar tools may benefit from conversational guidance that experts would find constraining; (3) the critique developed here applies most strongly to \emph{open-ended, multi-step analytical exploration} where the user must maintain and compare multiple hypotheses, states, and artifacts simultaneously.

The patterns proposed here are not anti-conversation but pro-hybrid: conversational scaffolding (like StatZ) and visual state externalization (like State Rails) can be combined. The 80/20 heuristic should be understood as task-dependent allocation, not a rigid prescription.

\subsection{Operationalizing the Framework}

The design patterns proposed here operationalize principles from Human-Centered AI \citep{shneiderman2022}: State Rails address transparency by externalizing hidden state; Deictic Interaction and Generative UI restore direct manipulation; the Pareto heuristic acknowledges that different tasks warrant different automation levels. Recent work on hybrid interfaces (DirectGPT's pointing-based interaction \citep{masson2024}, Google's generative UI showing strong user preference over markdown \citep{leviathan2025}) provides early empirical support for these patterns.

The sensemaking literature \citep{pirolli2005} distinguishes foraging (information gathering) from synthesis (structure creation). Chat excels at synthesis (summarizing, explaining, generating prose) but fails at foraging, which requires rapid filtering, comparison, and hypothesis testing. The hybrid patterns restore foraging capacity while preserving chat's synthesis strengths.

\textbf{Provenance and reproducibility.} A concern for Generative UI and Infinite Canvas is that ephemeral, dynamically-generated interfaces may lack the auditability of traditional dashboards. Future implementations should maintain provenance logs linking generated UI elements to the queries and data states that produced them, enabling replay, sharing, and verification across analysts.

\subsection{Implications for Practice}

For practitioners building LLM-powered analytical tools, the framework suggests concrete priorities: externalize state visibly (State Rail), support pointing as well as typing (Deictic Interaction), generate interactive artifacts rather than static text (Generative UI), and provide spatial persistence for analytical threads (Infinite Canvas). The 80/20 heuristic (chat for intent and synthesis, GUI for refinement and comparison) offers a practical allocation principle, modulated by task complexity and user expertise.

For researchers, the overload function (Equation~\ref{eq:overload}) provides testable predictions: cognitive load should scale with hidden state ($m - v$), errors should increase when $O > 0$, and bias amplification should correlate with serialization penalty $S(d)$. The hypotheses in Section~\ref{sec:agenda} are designed to be falsifiable through controlled experiments. Experimental designs should include conditions inspired by decomposition UIs to test whether structured, editable assumptions within chat containers can reduce $O$ without spatial externalization.

\subsection{The Skeuomorphic Moment}

We are likely at an inflection point analogous to the transition from command-line to graphical interfaces in the 1980s. Chat is the command line of AI: expressive and flexible for specification, but mismatched to spatial-visual tasks. The patterns described here represent early attempts to discover interface modalities native to AI-assisted analysis. Historical precedent suggests the dominant paradigm will not be chat with incremental improvements, but a qualitatively different interaction model.

%==============================================================================
\section{Limitations}
\label{sec:limitations}
%==============================================================================

This paper is primarily theoretical, synthesizing existing cognitive science research to generate predictions about interface design. Several limitations should be acknowledged.

\textbf{Model simplification.} The base overload model (Equation~\ref{eq:overload}) treats items as homogeneous, visibility as binary, and working memory capacity as fixed. Real systems require weighted items (filters vs. hypotheses contribute differently), graded visibility (pinned vs. minimized elements), and dynamic $W$ influenced by expertise, chunking, and external aids. The extended model (Equation~\ref{eq:extended}) sketches these extensions but remains untested. The exponential error mapping and linear serialization penalty are candidate functions, not derived results; alternative functional forms should be evaluated empirically.

\textbf{Empirical validation.} The cognitive load estimates and interaction time comparisons (Table~\ref{tab:time}) are derived from theoretical models (Fitts' Law, typing speed research) rather than direct experimental measurement of chat vs. hybrid interfaces for analytical tasks. The hypotheses in Section~\ref{sec:agenda} are falsifiable but have not yet been tested against modern chat-with-artifacts systems.

\textbf{Guided conversational systems.} As discussed in Section~\ref{sec:discussion}, well-scaffolded conversational systems that encode expert priors (e.g., StatZ) can reduce cognitive load relative to complex GUIs, particularly for novices. The critique developed here applies most strongly to open-ended, multi-step analytical exploration; the boundary conditions for when guided conversation suffices vs. when spatial externalization is necessary require empirical delineation.

\textbf{Intermediate designs.} Many modern ``chat'' tools are already hybrids (pinned charts, side panels, code cells). The framework's predictions are sensitive to where on the chat-to-hybrid continuum a system falls. Decomposition UIs that externalize editable assumptions within chat containers may achieve some benefits of state externalization without full GUI generation.

\textbf{Verbal overshadowing scope.} While the verbal overshadowing effect is well-established for face recognition, its application to data visualization is extrapolated. The effect is strongest under specific conditions (immediate verbalization, similar foils, holistic stimuli) that may or may not characterize typical analytical workflows. Tasks requiring fine-grained discrimination of similar distributions are predicted to show the strongest effects.

\textbf{Implementation challenges.} The hybrid patterns proposed here assume technical capabilities (real-time UI generation, modality switching, probabilistic rendering) that remain engineering challenges. The cognitive benefits are theoretical until working implementations can be evaluated.

\textbf{Individual differences.} Users vary in spatial ability, verbal fluency, and interface preferences. The framework does not yet account for how individual differences might moderate the Keyhole Effect or the effectiveness of hybrid patterns.

\textbf{Citation availability.} Some cited work (e.g., \citet{leviathan2025}) represents recent industry publications that may not be archived in traditional venues; their evidentiary status as peer-reviewed research should be understood accordingly.

%==============================================================================
\section{Conclusion}
%==============================================================================

Different cognitive tasks require different interface modalities. This is not opinion. It is architecture.

The brain has dedicated spatial processing systems centered on the hippocampus. It has separate verbal processing systems. It has working memory capacity limits around four chunks. It benefits from dual coding that engages both verbal and imagery systems. It extends into the environment through distributed cognition and cognitive offloading. Verbal overshadowing research shows that forcing visual information through linguistic channels does not merely fail to engage these systems; it can actively degrade the representations it touches.

Chat interfaces ignore all of this. They force spatial tasks through verbal channels. They provide no stable spatial structure for the visuospatial sketchpad. They prevent cognitive offloading by constantly changing. They offer no support for dual coding. They demand recall when recognition would suffice.

The Keyhole Effect, first identified by Woods four decades ago, takes on new urgency in the age of conversational AI. Analyzing multi-dimensional data through a chat window forces users to view large information spaces through a narrow, constantly-moving viewport. Spatial context, side-by-side comparison, and object permanence of previous findings are lost. Each scroll disrupts cognitive maps. Each typed refinement interrupts analytical flow.

The solution is hybrid interfaces where each modality handles what it does best. Chat for expressing complex, novel intent. Graphical interfaces for rapid, precise refinement. The eight patterns described here (Generative UI, Infinite Canvas, Deictic Interaction, State Rail, Ghost Layers, Mise en Place, Semantic Zoom, Probabilistic UI) provide specific mechanisms for making them work together.

The trajectory points toward interfaces that allocate interaction dynamically based on task demands: conversation for specification and synthesis, generated visual interfaces for exploration and comparison. The patterns and hypotheses presented here offer a research program for validating these claims empirically.

%==============================================================================
\bibliographystyle{plainnat}
\bibliography{keyhole_refs}

@article{shneiderman1983,
  author = {Ben Shneiderman},
  title = {Direct Manipulation: A Step Beyond Programming Languages},
  journal = {Computer},
  volume = {16},
  number = {8},
  pages = {57--69},
  year = {1983},
  publisher = {IEEE}
}

@book{shneiderman2022,
  author = {Ben Shneiderman},
  title = {Human-Centered AI},
  publisher = {Oxford University Press},
  year = {2022},
  address = {Oxford, UK}
}

@article{cowan2001,
  author = {Nelson Cowan},
  title = {The Magical Number 4 in Short-term Memory: A Reconsideration of Mental Storage Capacity},
  journal = {Behavioral and Brain Sciences},
  volume = {24},
  number = {1},
  pages = {87--114},
  year = {2001}
}

@article{okeefe1971,
  author = {John O'Keefe and Jonathan Dostrovsky},
  title = {The Hippocampus as a Spatial Map: Preliminary Evidence from Unit Activity in the Freely-moving Rat},
  journal = {Brain Research},
  volume = {34},
  number = {1},
  pages = {171--175},
  year = {1971}
}

@book{okeefe1978,
  author = {John O'Keefe and Lynn Nadel},
  title = {The Hippocampus as a Cognitive Map},
  publisher = {Oxford University Press},
  year = {1978}
}

@article{thompson1990,
  author = {Lonnie T. Thompson and Phillip J. Best},
  title = {Long-term Stability of the Place-field Activity of Single Units Recorded from the Dorsal Hippocampus of Freely Behaving Rats},
  journal = {Brain Research},
  volume = {509},
  number = {2},
  pages = {299--308},
  year = {1990}
}

@misc{nobel2014,
  author = {{Nobel Prize Committee}},
  title = {The {Nobel} Prize in Physiology or Medicine 2014},
  year = {2014},
  note = {John O'Keefe, May-Britt Moser, Edvard I. Moser}
}

@incollection{baddeley1974,
  author = {Alan D. Baddeley and Graham Hitch},
  title = {Working Memory},
  booktitle = {The Psychology of Learning and Motivation},
  editor = {Gordon H. Bower},
  volume = {8},
  pages = {47--89},
  publisher = {Academic Press},
  year = {1974}
}

@inproceedings{cockburn2004,
  author = {Andy Cockburn},
  title = {Revisiting 2D vs 3D Implications on Spatial Memory},
  booktitle = {Proceedings of the Fifth Conference on Australasian User Interface},
  volume = {28},
  pages = {25--31},
  year = {2004}
}

@article{bower1970,
  author = {Gordon H. Bower},
  title = {Analysis of a Mnemonic Device},
  journal = {American Scientist},
  volume = {58},
  number = {5},
  pages = {496--510},
  year = {1970}
}

@book{yates1966,
  author = {Frances A. Yates},
  title = {The Art of Memory},
  publisher = {University of Chicago Press},
  year = {1966}
}

@article{miller1956,
  author = {George A. Miller},
  title = {The Magical Number Seven, Plus or Minus Two: Some Limits on Our Capacity for Processing Information},
  journal = {Psychological Review},
  volume = {63},
  number = {2},
  pages = {81--97},
  year = {1956}
}

@article{logie1986,
  author = {Robert H. Logie},
  title = {Visuo-spatial Processing in Working Memory},
  journal = {Quarterly Journal of Experimental Psychology},
  volume = {38},
  number = {2},
  pages = {229--247},
  year = {1986}
}

@article{murdock1962,
  author = {Bennet B. Murdock},
  title = {The Serial Position Effect of Free Recall},
  journal = {Journal of Experimental Psychology},
  volume = {64},
  number = {5},
  pages = {482--488},
  year = {1962}
}

@book{paivio1971,
  author = {Allan Paivio},
  title = {Imagery and Verbal Processes},
  publisher = {Holt, Rinehart and Winston},
  year = {1971}
}

@book{paivio1986,
  author = {Allan Paivio},
  title = {Mental Representations: A Dual Coding Approach},
  publisher = {Oxford University Press},
  year = {1986}
}

@article{schooler1990,
  author = {Jonathan W. Schooler and Tonya Y. Engstler-Schooler},
  title = {Verbal Overshadowing of Visual Memories: Some Things are Better Left Unsaid},
  journal = {Cognitive Psychology},
  volume = {22},
  number = {1},
  pages = {36--71},
  year = {1990}
}

@article{alogna2014,
  author = {Alogna, V. K. and Attaya, M. K. and Aucoin, P. and Bahnik, S. and others},
  title = {Registered Replication Report: {Schooler} and {Engstler-Schooler} (1990)},
  journal = {Perspectives on Psychological Science},
  volume = {9},
  number = {5},
  pages = {556--578},
  year = {2014}
}

@article{meissner2001,
  author = {Christian A. Meissner and John C. Brigham},
  title = {A Meta-analysis of the Verbal Overshadowing Effect in Face Identification},
  journal = {Applied Cognitive Psychology},
  volume = {15},
  number = {6},
  pages = {603--616},
  year = {2001}
}

@book{hutchins1995,
  author = {Edwin Hutchins},
  title = {Cognition in the Wild},
  publisher = {MIT Press},
  year = {1995}
}

@article{kirshmaglio1994,
  author = {David Kirsh and Paul Maglio},
  title = {On Distinguishing Epistemic from Pragmatic Action},
  journal = {Cognitive Science},
  volume = {18},
  number = {4},
  pages = {513--549},
  year = {1994}
}

@article{hollan2000,
  author = {James Hollan and Edwin Hutchins and David Kirsh},
  title = {Distributed Cognition: Toward a New Foundation for Human-computer Interaction Research},
  journal = {ACM Transactions on Computer-Human Interaction},
  volume = {7},
  number = {2},
  pages = {174--196},
  year = {2000}
}

@article{fitts1954,
  author = {Paul M. Fitts},
  title = {The Information Capacity of the Human Motor System in Controlling the Amplitude of Movement},
  journal = {Journal of Experimental Psychology},
  volume = {47},
  number = {6},
  pages = {381--391},
  year = {1954}
}

@article{card1978,
  author = {Stuart K. Card and William K. English and Betty J. Burr},
  title = {Evaluation of Mouse, Rate-controlled Isometric Joystick, Step Keys, and Text Keys for Text Selection on a {CRT}},
  journal = {Ergonomics},
  volume = {21},
  number = {8},
  pages = {601--613},
  year = {1978}
}

@article{sweller1988,
  author = {John Sweller},
  title = {Cognitive Load During Problem Solving: Effects on Learning},
  journal = {Cognitive Science},
  volume = {12},
  number = {2},
  pages = {257--285},
  year = {1988}
}

@article{nguyen2022,
  author = {Quynh N. Nguyen and Anna Sidorova and Russell Torres},
  title = {User Interactions with Chatbot Interfaces vs. Menu-based Interfaces: An Empirical Study},
  journal = {Computers in Human Behavior},
  volume = {128},
  pages = {107093},
  year = {2022}
}

@article{tverskykahneman1974,
  author = {Amos Tversky and Daniel Kahneman},
  title = {Judgment Under Uncertainty: Heuristics and Biases},
  journal = {Science},
  volume = {185},
  number = {4157},
  pages = {1124--1131},
  year = {1974}
}

@article{wason1960,
  author = {Peter C. Wason},
  title = {On the Failure to Eliminate Hypotheses in a Conceptual Task},
  journal = {Quarterly Journal of Experimental Psychology},
  volume = {12},
  number = {3},
  pages = {129--140},
  year = {1960}
}

@article{rozenblitkeil2002,
  author = {Leonid Rozenblit and Frank Keil},
  title = {The Misunderstood Limits of Folk Science: An Illusion of Explanatory Depth},
  journal = {Cognitive Science},
  volume = {26},
  number = {5},
  pages = {521--562},
  year = {2002}
}

@article{rensink1997,
  author = {Ronald A. Rensink and J. Kevin O'Regan and James J. Clark},
  title = {To See or Not to See: The Need for Attention to Perceive Changes in Scenes},
  journal = {Psychological Science},
  volume = {8},
  number = {5},
  pages = {368--373},
  year = {1997}
}

@article{simonschabris1999,
  author = {Daniel J. Simons and Christopher F. Chabris},
  title = {Gorillas in Our Midst: Sustained Inattentional Blindness for Dynamic Events},
  journal = {Perception},
  volume = {28},
  number = {9},
  pages = {1059--1074},
  year = {1999}
}

@article{woods1984,
  author = {David D. Woods},
  title = {Visual Momentum: A Concept to Improve the Cognitive Coupling of Person and Computer},
  journal = {International Journal of Man-Machine Studies},
  volume = {21},
  number = {3},
  pages = {229--244},
  year = {1984}
}

@article{bennett2012,
  author = {Kevin B. Bennett and John M. Flach},
  title = {Visual Momentum Redux},
  journal = {International Journal of Human-Computer Studies},
  volume = {70},
  number = {6},
  pages = {399--414},
  year = {2012}
}

@inproceedings{masson2024,
  author = {Damien Masson and Sylvain Malacria and G{\'e}ry Casiez and Daniel Vogel},
  title = {{DirectGPT}: A Direct Manipulation Interface to Interact with Large Language Models},
  booktitle = {Proceedings of the 2024 CHI Conference on Human Factors in Computing Systems},
  year = {2024},
  publisher = {ACM},
  address = {New York, NY, USA}
}

@article{pirolli2005,
  author = {Peter Pirolli and Stuart Card},
  title = {The Sensemaking Process and Leverage Points for Analyst Technology as Identified Through Cognitive Task Analysis},
  journal = {Proceedings of International Conference on Intelligence Analysis},
  volume = {5},
  pages = {2--4},
  year = {2005}
}

@techreport{leviathan2025,
  author = {Yaniv Leviathan and Dani Valevski},
  title = {Generative {UI}: {LLMs} are Effective {UI} Generators},
  institution = {Google Research},
  year = {2025},
  note = {Available at generativeui.github.io}
}

\end{document}